%% file: z-main_file.tex
\documentclass[sigconf]{acmart}

\input{math_commands.tex}

\usepackage{hyperref}
\usepackage{url}
\usepackage{booktabs}
\usepackage{multirow}
\usepackage{float}
\usepackage[mathscr]{euscript}
\usepackage[ruled,vlined,linesnumbered]{algorithm2e}
\usepackage{graphicx}
\usepackage{subfigure}
\usepackage{balance}

\AtBeginDocument{%
  }

\copyrightyear{2026}
\acmYear{2026}
\setcopyright{cc}
\setcctype{by}
\acmConference[WSDM '26]{Proceedings of the Nineteenth ACM International Conference on Web Search and Data Mining}{February 22--26, 2026}{Boise, ID, USA}
\acmBooktitle{Proceedings of the Nineteenth ACM International Conference on Web Search and Data Mining (WSDM '26), February 22--26, 2026, Boise, ID, USA}
\acmDOI{10.1145/3773966.3777968}
\acmISBN{979-8-4007-2292-9/2026/02}

\begin{document}

\include{def}

\title{Multi-view Graph Condensation via Tensor Decomposition}

\author{Nicolas Roque dos Santos}
\affiliation{%
  \institution{University of California, Riverside}
  \city{Riverside}
  \state{California}
  \country{USA}
}
\email{nicolasr@ucr.edu}

\author{Dawon Ahn}
\affiliation{%
  \institution{University of California, Riverside}
  \city{Riverside}
  \state{California}
  \country{USA}
}
\email{dahn017@ucr.edu}

\author{Diego Minatel}
\affiliation{%
  \institution{University of São Paulo}
  \city{São Carlos}
  \state{São Paulo}
  \country{Brazil}
}
\email{dminatel@usp.br}

\author{Alneu de Andrade Lopes}
\affiliation{%
  \institution{University of São Paulo}
  \city{São Carlos}
  \state{São Paulo}
  \country{Brazil}
}
\email{alneu@icmc.usp.br}

\author{Evangelos Papalexakis}
\affiliation{%
  \institution{University of California, Riverside}
  \city{Riverside}
  \state{California}
  \country{USA}
}
\email{epapalex@cs.ucr.edu}

\renewcommand{\shortauthors}{Nicolas Roque dos Santos, Dawon Ahn, Diego Minatel, Alneu de Andrade Lopes, and Evangelos Papalexakis.}

\begin{abstract}
Graph Neural Networks (GNNs) have demonstrated remarkable results in various real-world applications, including drug discovery, object detection, social media analysis, recommender systems, and text classification. In contrast to their vast potential, training them on large-scale graphs presents significant computational challenges due to the resources required for their storage and processing. Graph Condensation has emerged as a promising solution to reduce these demands by learning a synthetic compact graph that preserves the essential information of the original one while maintaining the GNN's predictive performance. Despite their efficacy, current graph condensation approaches frequently rely on a computationally intensive bi-level optimization. Moreover, they fail to maintain a mapping between synthetic and original nodes, limiting the interpretability of the model's decisions. In this sense, a wide range of decomposition techniques have been applied to learn linear or multi-linear functions from graph data, offering a more transparent and less resource-intensive alternative. However, their applicability to graph condensation remains unexplored. This paper addresses this gap and proposes a novel method called Multi-view Graph Condensation via Tensor Decomposition (\method) to investigate to what extent such techniques can synthesize an informative smaller graph and achieve comparable downstream task performance. Extensive experiments on six real-world datasets demonstrate that \method effectively reduces graph size while preserving GNN performance, achieving up to a 4.0\% improvement in accuracy on three out of six datasets and competitive performance on large graphs compared to existing approaches. Our code is available at https://github.com/nicolasrsantos/gctd.
\end{abstract}

\ccsdesc[500]{Computing methodologies~Neural networks}
\ccsdesc[500]{Information systems~Data mining}
\ccsdesc[500]{Computing methodologies~Factorization methods}

\keywords{Graph Condensation; Tensor Decomposition; Graph Neural Networks; Matrix Tri-factorization}


\maketitle

\input{content/1-intro}
\label{sec:intro}

\input{content/2-related_work}
\label{sec:related_work}

\input{content/3-preliminaries}
\label{sec:preliminaries}

\input{content/4-single_view}
\label{sec:single_view}

\input{content/5-multi_view}
\label{sec:multi_view}

\input{content/6-exp}
\label{sec:exp}

\input{content/7-conclusion}
\label{sec:conclusion}


\begin{acks}
Research was supported in part by the National Science Foundation under CAREER grant no. IIS 2046086, grant no. 2431569 and CREST Center for Multidisciplinary Research Excellence in CyberPhysical Infrastructure Systems (MECIS) grant no. 2112650, and by the Agriculture and Food Research Initiative Competitive Grant no. 2020-69012-31914 from the USDA National Institute of Food and Agriculture. This study was also supported by the Coordination for the Improvement of Higher Education Personnel (CAPES) through the Institutional Internationalization Program (PRINT), Call No. 41/2017, and partially funded by CAPES (Finance Code 001), the São Paulo Research Foundation (FAPESP) [grants 20/09835-1, 22/02176-8, and 22/09091-8], and the National Council for Scientific and Technological Development (CNPq) [grants 303588/2022-5 and 406417/2022-9]. The views and conclusions contained in this document are those of the authors and should not be interpreted as representing the official policies, either expressed or implied, of the funding agencies. The U.S. Government is authorized to reproduce and distribute reprints for Government purposes not withstanding any copyright notation here on.
\end{acks}

\bibliographystyle{ACM-Reference-Format}
\balance
\bibliography{refs}

\end{document}

%% file: math_commands.tex
\usepackage{amsmath,amsfonts,bm}

\def\eqref#1{equation~\ref{#1}}
\def\Eqref#1{Equation~\ref{#1}}

\def\1{\bm{1}}

\DeclareMathAlphabet{\mathsfit}{\encodingdefault}{\sfdefault}{m}{sl}
\SetMathAlphabet{\mathsfit}{bold}{\encodingdefault}{\sfdefault}{bx}{n}

\def\gE{{\mathcal{E}}}

\def\gG{{\mathcal{G}}}

\def\gL{{\mathcal{L}}}

\def\gR{{\mathcal{R}}}
\def\gS{{\mathcal{S}}}
\def\gT{{\mathcal{T}}}

\def\gV{{\mathcal{V}}}

\def\gX{{\mathcal{X}}}

\newcommand{\R}{\mathbb{R}}



%% file: def.tex
\newcommand{\fix}{\marginpar{FIX}}
\newcommand{\new}{\marginpar{NEW}}
\newcommand{\method}{\textsc{GCTD}\xspace}
\newcommand{\nico}[1]{\textcolor{blue}{#1}}

\newcommand{\vagelis}[1]{{\color{red} vagelis: #1}}
\newcommand{\dawon}[1]{{\color{blue} dawon: #1}}
\newcommand{\blue}[1]{{\color{blue} #1}}
\newcommand{\red}[1]{{\color{red} #1}}
\newcommand{\orange}[1]{{\color{orange} #1}}
\newcommand{\green}[1]{{\color{teal} #1}}
\newcommand{\hide}[1]{}

\newcommand{\T}[1]{\boldsymbol{\mathscr{#1}}}
\newcommand{\tensor}[1]{\boldsymbol{\mathscr{#1}}}
\newcommand{\mat}[1]{\mathbf{#1}}
\newcommand{\vect}[1]{\mathbf{#1}}
\newcommand{\A}[1]{\mat{A}^{(#1)}}

%% file: content/1-intro.tex
\section{Introduction}
Graph is a ubiquitous data structure used to model a broad spectrum of real-world systems, encompassing transportation networks, protein-protein interactions, social media, and epidemic spreading~\cite{costa2011analyzing}. Over the past few years, Graph Neural Networks (GNNs) have become a pivotal tool for extracting representations from such data, supporting various downstream tasks. For example, they have been applied in the discovery of antibiotic compounds~\cite{halicin}, improving estimated time of arrival (ETA)~\cite{gmapsETA}, and detecting the spread of fake news on social networks~\cite{fake_news_gnn}. Despite their success, GNNs face significant scalability challenges when deployed on large-scale graphs~\cite{Hamilton2017GraphSAGE, gcond}. These issues become even more pronounced in scenarios like Neural Architecture Search or Continual Learning, where models require frequent retraining or incremental updates.

Numerous graph size reduction techniques have been proposed to mitigate these scaling issues~\cite{graph_reduction_survey}. Among these, condensation stands out as a promising one. Specifically, it aims to learn a smaller, representative graph that enables a GNN to achieve performance comparable to training on the original one. By reducing the graph, condensation significantly alleviates computational costs, making GNNs more feasible for large datasets. Existing methods typically generate condensed versions of the original graph using techniques such as Kernel Ridge Regression~\cite{kidd, NTK} or matching gradients~\cite{gcond}, training trajectories~\cite{sfgc}, eigenbasis~\cite{gdem}, and distribution~\cite{gcdm}. However, while such methods can compress graphs to an extreme degree without sacrificing much accuracy, they often rely on a complex bi-level optimization. Moreover, the need for multiple parameter initializations results in a costly triple-loop procedure. Additionally, these methods lack interpretability, losing the notion of how synthetic nodes relate to the original ones.

At the same time, there exists an extensive line of work in the literature that leverages matrix or tensor decomposition for graph-based tasks that learn simpler functions (\emph{i.e.}, linear or multi-linear) from data in contrast to the nonlinear ones learned by the GNNs~\cite{symmetric_nnmf_clustering, rolx, fusion_paper, pooling_td}. Furthermore, they offer a more transparent view of their decisions, providing a more interpretable approach. Importantly, the objectives of tensor decomposition and graph condensation are aligned: both aim to reduce the size and complexity of the original data while preserving its essential information, which in turn lowers the computational cost of downstream tasks. Nevertheless, to the best of our knowledge, no prior study has explored the application of tensor decomposition for synthesizing smaller graphs.

In this work, we address this gap by reframing the condensation task as a decomposition problem, investigating whether it can capture and transfer key information from a larger graph to a smaller one, thereby enabling a GNN to achieve performance comparable to that obtained when trained on the original graph. To achieve this, we propose a novel method called Multi-view \underline{G}raph \underline{C}ondensation via \underline{T}ensor \underline{D}ecomposition (\method). Specifically, we construct a multi-view graph by augmenting the original adjacency matrix into a third-order tensor through random edge perturbation. We then decompose this tensor and exploit the optimized latent factors to obtain a synthetic graph. The core idea is that the model discovers latent co-clusters within the data, enabling it to group nodes into synthetic ones based on shared patterns~\cite{smacd}.

Through extensive experiments on six real-world datasets, we demonstrate that \method effectively synthesizes smaller graphs, outperforming existing baselines on three out of six datasets with up to 4.0\% improvement in accuracy. Additionally, we show that using multi-view decomposition leads to better performance than its single counterpart.
Our contributions are summarized as follows:
\begin{itemize}
    \item We propose a novel method for graph condensation that leverages a tensor decomposition technique to reduce the original graph;
    \item We employ multi-view augmented graphs and show through a comprehensive analysis that it improves upon single-view decomposition;
    \item We conduct an extensive analysis to showcase to what extent decomposition methods can condense graphs;
\end{itemize}

%% file: content/2-related_work.tex
\section{Related Work}

\noindent \textbf{Graph Reduction.} Graph reduction encompasses many techniques developed to alleviate the computational cost of training graph-based algorithms by reducing the original graph into a smaller but informative one. Some examples of such methods include coreset~\cite{herding, k-center2}, sparsification~\cite{sparsification1, sparsification2}, and coarsening~\cite{coarseninghuang, coarseningloukas, dsaa}. Specifically, coreset methods extract representative samples that reflect the dataset. Furthermore, while sparsification reduces the original graph by removing nodes and edges based on properties such as eigenvalues, pairwise distances, or cuts, coarsening techniques merge nodes into supernodes based on their similarity~\cite{graph_reduction_survey, coarseninghuang, coarseningloukas}. Although these techniques can preserve information from the graph, they might not be sufficient for the downstream task. Additionally, they face a significant performance loss when the reduction rate is large (\emph{e.g.}, 99.9\%). Therefore, a new set of approaches known as condensation has emerged to tackle this problem. \\

\noindent \textbf{Graph Condensation.} Condensation is a technique that learns a small synthetic dataset that retains the essential patterns present in a larger one. Thus, when a model is trained on the reduced dataset, its performance is similar to when trained on the original one. These methods were originally developed for images ~\cite{dcg, dm} and later extended for graphs by~\citet{gcond}, which proposed GCond, a method that synthesizes a graph by matching the gradients of a GNN trained on both the original and simplified graphs. 


Since GCond, several strategies along the same line of research have emerged. For example, SGDD~\cite{sgdd} and GCSR~\cite{GCSR} incorporate the original structure information to generate the smaller graph. Conversely, SFGC~\cite{sfgc} and GEOM~\cite{geom} condense a graph by matching the training trajectories of the learned and original graphs. Rather than mimicking the training gradients or trajectories, GCDM~\cite{gcdm} and GDEM~\cite{gdem} synthesize a smaller graph through distribution and eigenbasis matching, respectively.

One of the major issues most graph condensation techniques face is the presence of a costly bi-level optimization. Therefore, DosCond~\cite{doscond} tackles this problem by performing a single gradient matching step to synthesize a reduced graph. Moreover, KiDD~\cite{kidd} and SNTK~\cite{NTK} avoid the bi-level matching by reframing condensation within the perspective of the Kernel Ridge Regression method.

Other approaches include CGC~\cite{cgc}, which proposes a fast, training-free condensation method; DisCo~\cite{disco}, which introduces a disentangled, GNN-free framework for graph condensation; and SimGC~\cite{simgc}, which employs an MLP alongside a heuristic to preserve the distribution of node features.

\noindent \textbf{Brief comparison.}
In contrast to existing graph condensation methods, our approach leverages tensor decomposition rather than GNN-based learning. Moreover, while prior methods typically depend on a computationally expensive triple-loop optimization procedure (discussed in the next section), our method removes this requirement entirely. Finally, unlike single-view approaches, our model exploits a multi-view graph representation, highlighting the benefits of incorporating augmentations into condensation.

%% file: content/3-preliminaries.tex
\section{Preliminaries}
In this section, we present the formal definition of graph condensation and then discuss a limitation with one of the main strategies adopted by prior work. Then, we describe the preliminaries on tensors and their decompositions.

\subsection{Graph Condensation}
A graph is denoted as $\gG = (\gV, \gE)$, where $\gV=\{v_1,\ldots,v_n\}$ comprises a set of $N$ nodes and $\gE$ is the edge set. The graph's structure can be represented as an adjacency matrix $\mathbf{A} \in \R^{N \times N}$, where each entry $A_{ij}=1$ if there is an edge between nodes $v_i$ and $v_j$, and $0$ otherwise. Moreover, $\mathbf{X} \in \R^{N \times d}$ represents the $d$-dimensional node feature matrix and $\mathbf{Y} \in \{0, \ldots, C-1\}^N$ are the node labels over $C$ classes.

Given a target graph $\gG^{\gT}=\{\mathbf{A}^{\gT},\mathbf{X}^{\gT},\mathbf{Y}^{\gT}\}$, the goal of graph condensation is to reduce $\gG^{\gT}$ into a smaller graph $\gG^{\gS}=\{\mathbf{A}^{\gS},\mathbf{X}^{\gS},\mathbf{Y}^{\gS}\}$, where $\mathbf{A}^{\gS} \in \R^{N' \times N'}$, $\mathbf{X}^{\gS} \in \R^{N' \times d}$, $\mathbf{Y}^{\gS} \in \{0, \ldots, C-1\}^{N'}$, and $N' \ll N$, such that a GNN trained on $\gG^{\gS}$ achieves similar performance to when trained on the significantly larger graph $\gG^{\gT}$. Existing work for graph condensation approaches this task as a bi-level optimization problem~\cite{gcond, sfgc, sgdd}. Formally, it is defined as:
\begin{align}\label{eq:bi_level}
\begin{gathered}
    \min_{\gG^{\gS}} \gL (\text{GNN}_{\boldsymbol{\theta}_{\gG^{\gS}}}(\mathbf{A}^\gT,\mathbf{X}^\gT), \mathbf{Y}^\gT)\\
    \text { s.t. } \quad \boldsymbol{\theta}_{\gG^{\gS}}=\underset{\boldsymbol{\theta}}{\arg \min } \gL(\text{GNN}_{\boldsymbol{\theta}}({\bf A^{\gS}},{\bf X^{\gS}}), {\bf Y^{\gS}}).
\end{gathered}
\end{align}

Here, the outer loop is responsible for optimizing the synthetic graph based on, for example, a gradient matching loss, while the inner loop trains a GNN parameterized by $\boldsymbol{\theta}$ on the synthetic dataset. Furthermore, this double loop is computed multiple times with different parameter initializations to ensure the entire process is not overfitting to a specific $\boldsymbol{\theta}$. Consequently, it becomes a triple-loop procedure, making graph condensation models that follow this scheme computationally costly in terms of time, storage, and processing power. In addition to the aforementioned expensive step, most methods leverage a GNN to learn a condensed graph. This results in a black-box model, making it difficult to establish a clear connection between the original and synthetic nodes. 

\subsection{Tensors}
\noindent \textbf{Notations.}
Tensors are multi-dimensional arrays that generalize one-dimensional arrays (or vectors) and two-dimensional arrays (or matrices) to higher dimensions. 
Throughout this paper, we use boldface Euler script letters (\emph{e.g.}, $\T{X}$) to denote tensors, boldface capitals (e.g., $\mat{A}$) to denote matrices, boldface lowercases (\emph{e.g.}, $\vect{a}$) to denote vectors, and unbolded letters (\emph{e.g.}, $A$, $a$) to denote scalars and coefficients. We refer to a tensor's dimension as its \emph{mode} or \emph{order}. The \emph{slices} are two-dimensional sections of a tensor, defined by fixing all but two indices. For example, the $k$-th frontal slice of a third-order tensor $\T{X} \in \mathbb{R}^{I \times J \times K}$ is a matrix denoted as $\mat{X}_{k} \in \mathbb{R}^{I \times J}$. Moreover, the $n$-mode product defines the multiplication of a tensor with a matrix in mode $n$.
For example, the $n$-mode product of a tensor $\T{X} \in \mathbb{R}^{I \times J \times K}$ with a matrix $\mat{U} \in \mathbb{R}^{R \times I}$ along the first mode is denoted by $\T{X} \times_1 \mat{U} (\in \mathbb{R}^{R \times J \times K})$. 
 \\

\noindent \textbf{Tensor decomposition.} 
Tensor decomposition techniques aim to decompose tensors into low-rank latent components, facilitating data mining and analysis. Among the most widely used models are CANDECOMP/PARAFAC (CP) and Tucker decompositions, both of which have seen extensive development and application across a diverse range of fields~\cite{tensor_survey,rabanser2017introduction}. In this work, we leverage a variant of Tucker called RESCAL~\cite{rescal} and, thus, we describe them next.

Tucker decomposition approximates a given third mode tensor $\T{X} \in \mathbb{R}^{I \times J \times K}$ as follows:
\begin{equation} \label{eq:tucker}
 \T{X} \approx \T{R} \times_1 \mat{U}^{(1)} \times_2 \mat{U}^{(2)}  \times_3 \mat{U}^{(3)} ,
\end{equation}
\noindent where $\times_1, \times_2$, and $\times_3$ denote the $n$-mode product along the first, second, and third modes, respectively. Factor matrices $\mat{U}^{(1)} \in \mathbb{R}^{I \times R_1}, \mat{U}^{(2)} \in \mathbb{R}^{J \times R_2}$ and $\mat{U}^{(3)} \in \mathbb{R}^{K \times R_3}$ are considered as the principal components in each mode. The core tensor $\T{R} \in \mathbb{R}^{R_1 \times R_2 \times R_3}$ captures the interaction between each component. Here, $R_1, R_2$, and $R_3$ denote the number of components in each mode and are smaller than $I, J, K$, respectively. Furthermore, it has been demonstrated that $\T{R}$ can be viewed as a compressed version of 
$\T{X}$~\cite{kolda2009tensor}.

RESCAL was initially proposed for relational learning and is particularly useful when the frontal slices of the given tensors exhibit symmetry. It is a special case of the Tucker decomposition, as described in \Eqref{eq:tucker}, with a few key differences. Specifically, the core tensor $\T{R} \in \mathbb{R}^{R_1 \times R_2 \times K}$ is employed, and the factor matrix $\mat{U}^{(3)}$ is set as an identity matrix. Additionally, the factor matrices $\mat{U}^{(1)}$ and $\mat{U}^{(2)}$ are identical and denoted by $\mat{U}$, as follows:
\begin{equation} \label{eq:tucker:rescal}
 \T{X} \approx \T{R} \times_1 \mat{U} \times_2 \mat{U} 
 \Leftrightarrow  \mat{X}_{k} \approx \mat{U}  \mat{R}_k \mat{U}^\top,
\end{equation}
\noindent where $\mat{R}_k \in \mathbb{R}^{R_1 \times R_2}$ indicates the relations between latent components. It is important to highlight that RESCAL does not compress the third mode of the tensor.


%% file: content/4-single_view.tex
\section{First attempt: Single-view Graph Modeling}
Motivated by prior work employing Matrix Factorization (MF) for tasks such as clustering~\cite{mtf_clustering, doc_clustering}, dimensionality reduction~\cite{structure_nmf, discriminative_dr}, and compression~\cite{nn_compression, gradzip}, all of which aim to simplify data complexity, our initial approach explored single-view graph condensation by factorizing the adjacency matrix $\mathbf{A}^\gT$ of the graph $\gG^\gT$.

In this work, we use a variant of MF called Matrix Tri-factorization (MTF), which decomposes a matrix into the product of three lower-dimensional factors~\cite{mtf_clustering}. Specifically, we approximate $\mathbf{A}^{\gT} \approx \mathbf{U} \mathbf{R} \mathbf{V}^{\top}$, where $\mathbf{U}$ and $\mathbf{V}$ capture the row and column spaces of $\mathbf{A}^{\gT}$, while $\mathbf{R}$ encodes the interactions between them. However, given the nature of the undirected graphs, we adopt a symmetric formulation of MTF where $\mathbf{V}=\mathbf{U}$, yielding the following reconstruction:
\begin{equation}
    \label{eq:single_view_decomp}
    \mathbf{A}^{\gT} \approx \mathbf{U} \mathbf{R} \mathbf{U}^{\top},
\end{equation}
\noindent where $\mathbf{U} \in \R^{N \times N'}$ and $\mathbf{R} \in \R^{N' \times N'}$. Here, $N$ is the number of nodes in the original graph, $N'= rN$ is the number of nodes in the condensed graph, $r$ is the condensation ratio, and $N' \ll N$. Note that if the core tensor in ~\Eqref{eq:tucker:rescal} is a matrix, ~\Eqref{eq:single_view_decomp} is equivalent to the ~\Eqref{eq:tucker:rescal}.\\

\noindent \textbf{Reconstruction objective.} After computing $\mathbf{A}^\gT$, we optimize both $\mathbf{U}$ and $\mathbf{R}$ using gradient descent, employing mean squared error (MSE) as the reconstruction loss function. Additionally, we identified empirically that normalizing the MSE by the sum of the squared elements of the original matrix $\mathbf{A}$ enhances convergence speed and stability. Therefore, the final loss function is defined as:
\begin{equation}
    \label{eq:loss_fn}
    \mathcal{L}_{rec} = \frac{\frac{1}{n} \sum_{i,j} \left(a_{ij} - \hat{a}_{ij}\right)^2}{\sum_{ij}a_{ij}^2},
\end{equation}
\noindent where $a_{ij}$ and $\hat{a}_{ij}$ represent elements from the original and reconstructed matrices, respectively.\\

\noindent \textbf{Uncovering the condensed graph.} The reconstruction from~\autoref{eq:single_view_decomp} provides a low-rank approximation of the original adjacency matrix, grouping together nodes with similar patterns and capturing their relationships through $\mathbf{U}$ and $\mathbf{R}$, respectively. The interaction between the rows of $\mathbf{U}$, encoded by $\mathbf{R}$, are then used as the adjacency matrix $\mathbf{A}^\gS$ of the condensed graph.

In order to extract the synthetic nodes from this decomposition, we apply the K-Means\footnote{We use the implementation provided by the Faiss library, available at faiss.ai.} algorithm to the rows of the factor matrix $\mathbf{U}$. Formally, we minimize the following error function:
\begin{align}\label{eq:kmeans}
\begin{gathered}
    \mathcal{L}_{\text{K-Means}} = \sum_{i=1}^{I} \sum_{j=1}^{J} \delta_{ij} \lVert \mathbf{u}_i - \mu_j \rVert^2 \\
    \text { s.t } \quad \delta_{ij} = \begin{cases} 1, \text{if } j = \arg \min_k \lVert \mathbf{u}_i - \mu_k \rVert^2  \\ 0, \text{otherwise} \end{cases}
\end{gathered}
\end{align}
\noindent where $\mathbf{u}_i$ is a row from the factor matrix, $\mu_j$ corresponds to a cluster centroid, and $\delta_{ij} \in \{0,1\}$ indicates to which of the $J$ clusters $\mathbf{u}_i$ is assigned to. We initialize the centroids randomly and the number of clusters is set to match the number of columns of $\mathbf{U}$, denoted $N'$, which corresponds to the size of $\gG^{\gS}$. This process determines which nodes from the original graph $\gG^{\gT}$ should be grouped together in the reduced graph, providing the necessary information to assign class labels, splits, and features to the synthetic nodes.

Given the assignments for cluster $j$, the split of the corresponding synthetic node is chosen as the most frequent split among the original nodes $i$ for which $\delta_{i,j}=1$. The class label is then determined by considering only the nodes from that split. For example, if five nodes are assigned to cluster $j$ with splits {train, val, train, test, train} and classes {0, 2, 1, 0, 0}, the synthetic node will be assigned to the training split and to class 0, which is the most frequent class among the training nodes. Finally, the features of the synthetic node are computed by averaging the embeddings of the nodes assigned to it that share both its split and class. Throughout this process, we prioritize underrepresented splits and classes to ensure the condensed graph preserves the original distribution.\\


\noindent \textbf{Remarks on single-view decomposition.}
While single-view decomposition enables the synthesis of a condensed graph, we extend this approach to a multi-view setting, as recent studies show that multi-view augmentation improves model performance, generalization, and robustness~\cite{tong_aug_paper, multiview_paper, graph_aug_survey}. Moreover, as demonstrated in Section~\ref{sec:experimental_results}, this strategy yields notable gains in GNN performance on condensed graphs compared to the single-view case. Thus, in the following section, we introduce our method, which leverages tensor decomposition to condense multi-view augmented graphs.


%% file: content/5-multi_view.tex
\section{Proposed Method}
\begin{figure*}[ht]
    \centering
    \hspace{-4mm}
    \includegraphics[width=\linewidth]{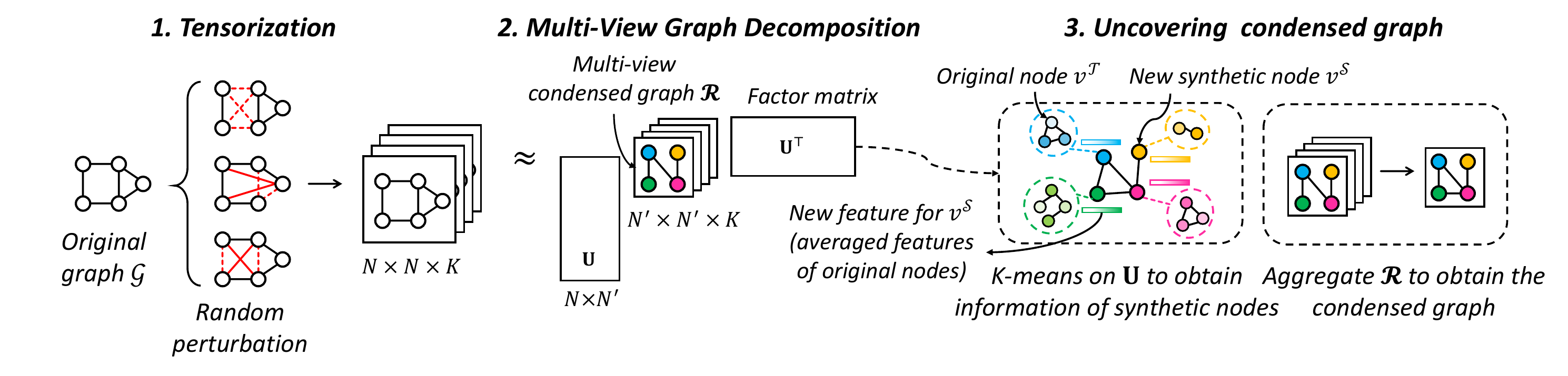}
    \Description{Pipeline of \method. We construct a tensor by augmenting the graph's adjacency matrix $\mathbf{A}$ and stacking them together in the third dimension with $\mathbf{A}$. Then, we apply non-negative RESCAL to the given tensor to extract low-rank structures $\mat{U}$ and a multi-view condensed graph $\T{R}$.
    Lastly, we obtain a condensed graph by aggregating $\T{R}$ along the third mode, and we identify a feature and label for each synthetic node by applying K-Means to $\mat{U}$}
    \vspace{-0.1in}
    \caption{
    Pipeline of \method. We construct a tensor by augmenting the graph's adjacency matrix $\mathbf{A}^\gT$ and stacking them together in the third dimension with $\mathbf{A}^\gT$. Then, we apply non-negative RESCAL to the given tensor to extract low-rank structures $\mat{U}$ and a multi-view condensed graph $\T{R}$.
    Lastly, we obtain a condensed graph by aggregating $\T{R}$ along the third mode, and we compute the feature and label for each synthetic node by applying K-Means to $\mat{U}$.
    }
    \label{fig:pipeline}
\end{figure*}

We introduce our proposed method, \method. We begin by outlining the process for obtaining multi-view graphs, followed by a detailed explanation of the decomposition procedure. Afterward, we describe the steps involved in deriving the condensed graph. Figure~\ref{fig:pipeline} illustrates the complete pipeline of our method.\\

\noindent \textbf{Multi-view augmentation.}\label{sec:multi_view_aug}
Data augmentation (DA) methods have been widely explored across various domains, including Graph Machine Learning, Computer Vision, and Natural Language Processing~\cite{tong_aug_paper, cv_aug, nlp_aug}. They aim to generate variations of existing data, thereby alleviating the burden of collecting and labeling additional samples. Existing work has demonstrated that synthetic samples produced through DA enhance generalization, improving downstream task performance~\cite{autoaugment, aug_diffusion, dropedge}. Furthermore, the variations created by DA enable models to effectively learn to distinguish signal from noise~\cite{tong_aug_paper}.

The utilization of DA to generate multiple views of a dataset and jointly extract information from them has been investigated in several studies~\cite{simclr, graphcl}. Additionally, it has also been shown that employing a multi-view graph not only improves robustness against adversarial attacks but also reduces the variance of the GNNs' predictions on augmented graphs~\cite{multiview_paper}. 

Motivated by the aforementioned work on augmentation and multi-view graphs, we generate multiple views of the graph by augmenting its adjacency matrix $\mathbf{A}^\gT$ to obtain perturbed versions of its original structure. Specifically, we construct a set of $K$ matrices, $S = \{s_1, \ldots, s_K\}$, where each matrix represents an augmented version of $\mathbf{A}^\gT$ with edges randomly removed and added based on prespecified probabilities.

For each perturbed matrix $s_k$, edges are first dropped randomly with probability $p_r$. This results in a reduced edge count of $M' = M(1 - p_r)$, where $M$ is the original number of edges. Subsequently, new edges are added randomly with probability $p_a$, drawn from the set of potential edges not already present in the graph. The total number of edges after this step is updated to $M' = M' + p_a \cdot \left( \frac{N(N-1)}{2} - M'\right)$, where $\frac{N(N-1)}{2}$ represents the total number of possible edges in an undirected graph with $N$ nodes. In practice, we set $p_r$ and $p_a$ to small values (i.e., $0.05 \leq p_r, p_a \leq 0.2$) to ensure that the resulting perturbed graphs remain with a similar size to the original one, with $M' \approx M$ on average.

 The perturbed matrices are then combined to form a tensor $\T{X}$, where the first slice corresponds to $\mathbf{A}^\gT$, and the remaining ones consist of the $K$ matrices in $S$. The final structure of $\T{X}$ is illustrated in Step 1 of Figure~\ref{fig:pipeline}. It is important to note that, to mitigate computational overhead during the augmentation process, all matrices in $S$ are precomputed.\\

\noindent \textbf{Decomposing a multi-view graph.}
We reconstruct the multi-view graph $\T{X}$ using the tensor decomposition RESCAL~\cite{rescal}. Formally, we compute the following operation for each slice $k$ of $\T{X}$:
\begin{equation}
    \label{eq:rescal}
    \tilde{\mat{X}}_k = \mathbf{U} \mat{R}_k \mathbf{U}^\top,
\end{equation}
\noindent where $\mathbf{U} \in \R^{N \times N'}$ is the factor matrix capturing latent components in the given graph and $\mat{R}_{k}$ is the $k$-th frontal slice of the core tensor $\T{R} \in \R^{N' \times N' \times K}$, indicating relations between latent components existing in the $k$-th view of the graph.
Note that $N'$ is the size of the condensed graph, and $K$ is the total number of augmented graphs.

The formulation presented in Equation~\ref{eq:rescal} consists of an operation that produces $k$ dense matrices, which is computationally expensive for large graphs. To make things worse, keeping them in memory during runtime is resource-intensive since we have to store multiple $N \times N$ matrices. To address this issue, we instead adopt the sparse version of the decomposition where we reconstruct only the observed entries of the original tensor as follows:
\begin{equation}
    \label{eq:sparse_rescal}
    \tilde{x}_{ijk} = \mathbf{u}_i^\top \mat{R}_{k} \mathbf{u}_j.
\end{equation}
Since this operation involves only observed entries (\emph{i.e.}, nonzero values), providing negative samples to the model is essential to prevent overfitting to the training data. Therefore, we randomly generate negative examples in 1:1 proportion to the available nonzero values, ensuring a balanced representation of both positive and negative samples. It is noteworthy that this process is precomputed to minimize the computational overhead during the decomposition. In addition to performing the sparse operation on the nonzero values, we adopt mini-batching in our model's reconstruction step to further reduce the memory footprints. 

After reconstruction, we optimize $\T{R}$ and $\mathbf{U}$ using the same objective as in the single-view decomposition, with a slight modification to Equation~\ref{eq:loss_fn} to account for tensor entries:
\begin{equation}
    \label{eq:loss_fn_tensor}
    \mathcal{L}_{rec} = \frac{\frac{1}{n} \sum_{i,j,k} \left(x_{ijk} - \tilde{x}_{ijk}\right)^2}{\sum_{i,j,k} x_{ijk}^2},
\end{equation}
\noindent where $x_{ijk}$ and $\tilde{x}_{ijk}$ are entries from the original and reconstructed tensor, respectively.\\

\noindent \textbf{Nonnegativity and sparsity constraint.} 
Given the nonnegative nature of most real-world graphs, we impose a constraint in our method to maintain the nonnegative characteristics of the data. It is important to note that we introduce hard constraints instead of soft regularization-based ones that prior work did~\cite{rescal_constrained}. Specifically, we first ensure that the random initialization of the factor matrix and core tensor comprises only nonnegative values by applying the absolute function to them (\emph{i.e.}, $\mathbf{U} \leftarrow \lvert \mathbf{U} \rvert$ and $\T{R} \leftarrow \left| \T{R} \right|$). Additionally, since every optimization step of \method can shift some values of $\mathbf{U}$ and $\T{R}$ to negative, we apply the nonlinear activation function ReLU to them in every decomposition epoch to ensure they remain nonnegative, as shown in Algorithm~\ref{alg:gctd}. Besides guaranteeing the constraint mentioned above, applying ReLU inherently induces sparsity in the learned graph since it sets a portion of the values to zero, reducing the storage necessary for the condensed graph.\\

\noindent \textbf{Computing the structure and features of a condensed graph.}
While it is not possible to make uniqueness arguments for \method similar to those for CP~\cite{tensor_survey}, by imposing nonnegativity and sparsity to $\T{R}$ and $\mathbf{U}$, we are constraining the otherwise non-unique Tucker-like model to a smaller space of solutions~\cite{nmf_uniqueness}. These imposed constraints guide \method toward behaving close to a constrained CP model, enabling it to uncover latent co-clusters from the data effectively~\cite{smacd}. Thus, we can interpret that the computed factor $\mathbf{U}$ co-clusters the original nodes based on similar patterns they share, and the core tensor $\T{R}$ captures the relations between them. 

We specifically uncover the condensed graph from $\T{R}$ and $\mathbf{U}$ similarly to the process we described in the single-view decomposition approach. The only existing difference is that a tensor is the latent component that captures the interactions between the entities in the factors, instead of a matrix. Therefore, after deriving the information for each synthetic node, we compute the structure of the condensed graph $\gG^{\gS}$. This is done by averaging each $\T{R}_{i,j,:}$ from the core tensor to form the adjacency matrix $\mathbf{A}^{\gS}$. Since the aggregation process of the core tensor may result in a non-symmetric matrix, we add the corresponding off-diagonal values to symmetrize $\mathbf{A}^{\gS}$.\\

\noindent \textbf{Algorithm.} We summarize the step-by-step of \method in Algorithm~\ref{alg:gctd}. Our method takes as input the observed entries of the multi-view graph along with the negative samples. The factor matrix and core tensor are randomly initialized, and the absolute function is applied to ensure their nonnegativity. From lines 6 to 9 of the algorithm, the adjacency matrix is decomposed and the latent components are optimized while maintaining their nonnegativity throughout the process. Convergence is determined either after 200 epochs or when the variation in reconstruction loss between consecutive steps falls below $10^{-7}$, as previously described. Finally, the condensed graph is uncovered by averaging the core tensor and applying KMeans clustering to the factor matrix, extracting the necessary information to generate the features, classes, and splits.

\begin{algorithm}[!htp]
\SetAlgoVlined
\small
\textbf{Input:} Pre-computed multi-view graph $\gX$ with negative samples.\\
\textbf{Output:} Condensed graph $\gG^{\gS}=(\mathbf{A}^{\gS}, \mathbf{X}^{\gS}, \mathbf{Y}^{\gS})$\\
Initialize matrix $\mathbf{U}$ and core tensor $\gR$ randomly.\\
$\mathbf{U} \gets \lvert \mathbf{U} \rvert$; $\gR \gets \lvert \gR \rvert$ \\
\While{$\text{convergence is not achieved}$}{ 
    Reconstruct $\T{X}$ according to Equation~\ref{eq:sparse_rescal}.\\
    Compute the reconstruction error $\mathcal{L}_{\text{rec}}$ according to Equation~\ref{eq:loss_fn_tensor}.\\
    Update $\mathbf{U} \gets \mathbf{U}-\eta\nabla_{\mathbf{U}}\mathcal{L}_{\text{rec}}$ and $\T{R} \gets \T{R}-\eta\nabla_{\T{R}}\mathcal{L}_{\text{rec}}$\\
    $\mathbf{U} \gets \text{ReLU}\left( \mathbf{U} \right)$; $\T{R} \gets \text{ReLU}\left( \T{R} \right)$\\
}
Compute $\mathbf{A}_{ij}^{\gS} \gets \text{Average}\left( \T{R}_{i,j,:} \right), \quad \forall \, i, j$\\
Cluster $\mathbf{U}$ rows according to Equation~\ref{eq:kmeans}.\\
Average embeddings of the nodes assigned by K-Means to get $\mathbf{X}^{\gS}$.\\
Assign to each synthetic node the dominant split and class among its assigned nodes, focusing on underrepresented classes first.\\
\caption{Graph Condensation via Tensor Decomposition (\method)}
\label{alg:gctd}
\end{algorithm}

The split of a synthetic node is chosen based on the most frequent split among its assigned nodes, with preference given to underrepresented splits according to the target distribution. To assign a class label, we first filter the assigned nodes to those within the selected split and examine their class labels. Instead of simple majority voting, we compute the class frequencies and prioritize underrepresented classes. The first class whose current count is below its target proportion is selected; if all are satisfied, a class is chosen randomly among the candidates. This strategy helps maintain class balance in the condensed graph. Finally, the feature representation of the synthetic node is computed by averaging the embeddings of its assigned nodes.\\

\noindent \textbf{Complexity Analysis.} \method's time complexity consists of three components: tensor decomposition, clustering, and adjacency matrix generation. For tensor decomposition, operations are performed only on observed entries (Equation~\ref{eq:sparse_rescal}), with each entry requiring $\mathcal{O}(N'^2)$ time. With approximately $M$ observed entries per view and $K$ views ($2 \leq K \leq 5$), the decomposition cost is $\mathcal{O}(N'^2M K)$. For the sparse graphs typically used in graph condensation, $M \ll N^2$ and $N' \ll N$. KMeans clustering incurs $\mathcal{O}(T N'^2 N)$ time, where $T$ denotes the number of iterations ($T=20$), and constructing the condensed adjacency matrix by averaging along the third mode adds $\mathcal{O}(N'^2)$. Thus, the overall complexity is $\mathcal{O}(N'^2 M K) + \mathcal{O}(T N'^2 N) + \mathcal{O}(N'^2)$, which remains practical since graph condensation explicitly targets small $N'$.

The space complexity is dominated by the factor matrix $\mathbf{U}$ and the core tensor $\T{R}$, requiring $\mathcal{O}(N \times N')$ and $\mathcal{O}(N'^2 \times K)$ memory, respectively. Since only observed entries are stored, the original graph requires $\mathcal{O}(M)$ space, while the multi-view augmented graphs require $\mathcal{O}(M')$. In practice, $M' \approx M$, as the augmentation probabilities $p_r$ and $p_a$ are set between 0.05 and 0.2.




%% file: content/6-exp.tex
\section{Experiments}
In this section, we present the experimental results of our proposed method for graph condensation, \method. 

\subsection{Experimental Settings}

\noindent\textbf{Baselines.}
We compared our condensation method against twelve baselines, which include three graph coreset methods (Random, Herding~\cite{herding}, and K-Center~\cite{k-center2}), one coarsening method~\cite{coarseninghuang}, two graph distillation methods: SGDD~\cite{sgdd}, and GDEM~\cite{gdem}, and six graph condensation methods: GCond~\cite{gcond}, SFGC~\cite{sfgc}, SNTK~\cite{NTK}, GCSR~\cite{GCSR}, GCDM~\cite{gcdm}, and CGC~\cite{cgc}.

It is worth noting that although graph coarsening techniques~\cite{coarseninghuang, coarseningloukas, bernnet, convmatching} are conceptually related to graph condensation, numerous studies have consistently shown that coarsening tends to underperform condensation methods on downstream tasks at high condensation ratios~\cite{gcond, sgdd, geom, gdem, cgc}. Consequently, our evaluation focuses on distillation and condensation baselines, which is consistent with the empirical comparison practices adopted in the majority of recent works in this area.\\

\noindent\textbf{Datasets.}
Following prior work in graph condensation~\cite{gcond, sgdd, sfgc}, we evaluate our method on six datasets that range from a few thousand to hundreds of thousands of nodes: four transductive graphs (Cora, Citeseer, Pubmed, and Ogbn-arxiv) and two inductive graphs (Flickr and Reddit). Cora, Citeseer, and Pubmed are taken from the PyTorch Geometric library\footnote{\url{https://pytorch-geometric.readthedocs.io/}}, Flickr and Reddit from GraphSAINT \cite{GraphSAINT}, and Ogbn-arxiv from the Open Graph Benchmark \cite{ogbdatasets}. To maintain consistency with prior work, all experiments utilize the official public splits for each dataset. Furthermore, we adopt the same dataset-specific condensation ratios employed in previous studies to enable a direct and fair comparison. A summary of dataset characteristics can be found in Table \ref{tab:datasets}.
\begin{table}[!htp]
\caption{Statistics of the datasets. The first four datasets are transductive, while the last two are inductive.}
\label{tab:datasets}
\setlength{\tabcolsep}{4pt}
\footnotesize                              
\centering
\begin{tabular}{lccccc}
\toprule
\textbf{Dataset} & \textbf{Nodes} & \textbf{Edges} & \textbf{Classes} & \textbf{Features} & \textbf{Train/Val/Test} \\ \midrule
Cora             & 2,708            & 5,429            & 7                  & 1433                & 140/500/1000            \\
Citeseer         & 3,327            & 4,732            & 6                  & 3,703               & 120/500/1000            \\
Pubmed           & 19,717           & 44,338           & 3                  & 500                 & 50/500/1000             \\
Ogbn-arxiv       & 169,343          & 1,166,243        & 40                 & 128                 & 90,941/29,799/48,603    \\ \midrule
Flickr           & 89,250           & 899,756          & 7                  & 500                 & 44,625/22,312/22,313    \\
Reddit           & 232,965          & 57,307,946       & 210                & 602                 & 153,932/23,699/55,334   \\ \bottomrule
\end{tabular}
\end{table}

\begin{table*}[htbp]
\caption{Node classification accuracy (\%) of the baselines and our proposed method on six datasets with three different condensation ratios. We report the average of ten runs and the corresponding standard deviation. The best and second-best results are in bold and underlined, respectively. OOM stands for out of memory, which in our case was 49GB.}\vspace{-0.05in}
\label{tab:results}
\setlength{\tabcolsep}{1.5pt}
\centering
\small
\begin{tabular}{lcccccccccccccccc}
\toprule
\multirow{2}{*}{Datasets} & \multirow{2}{*}{Ratio (\%)} &
\multicolumn{4}{c}{Traditional Methods} &
\multicolumn{9}{c}{Condensation and Distillation Methods} &
\multirow{2}{*}{Full} \\
\cmidrule(lr){3-6}\cmidrule(lr){7-15}
 & & Random & Herding & K-Center & Coarse &
 GCond & SFGC & SGDD & SNTK & GCSR & GCDM & GDEM & CGC & \textbf{GCTD} & Dataset \\ \midrule
\multirow{3}{*}{Cora}
 & 1.3 & 63.6$\pm$3.7 & 67.0$\pm$1.3 & 64.0$\pm$2.3 & 31.2$\pm$0.2 & 79.8$\pm$1.3 & 77.7$\pm$1.8 & 79.1$\pm$1.3 & 78.4$\pm$1.4 & 79.9$\pm$0.7 & 79.1$\pm$0.9 & 68.0$\pm$0.1 & \textbf{82.6$\pm$0.3} & \underline{81.4$\pm$1.6} & \multirow{3}{*}{81.4$\pm$0.6} \\
 & 2.6 & 72.8$\pm$1.1 & 73.4$\pm$1.0 & 73.2$\pm$1.2 & 65.2$\pm$0.6 & 80.1$\pm$0.6 & 79.3$\pm$0.8 & 79.0$\pm$1.9 & 79.7$\pm$0.9 & 80.6$\pm$0.8 & 80.5$\pm$0.3 & 72.8$\pm$0.8 & \underline{81.2$\pm$0.6} & \textbf{84.0$\pm$0.4} &  \\
 & 5.2 & 76.8$\pm$0.1 & 76.8$\pm$0.1 & 76.7$\pm$0.1 & 70.6$\pm$0.1 & 79.3$\pm$0.3 & 79.4$\pm$0.5 & 80.2$\pm$0.8 & 80.5$\pm$0.6 & \underline{81.2$\pm$0.9} & 80.2$\pm$0.5 & 77.4$\pm$0.6 & \textbf{82.1$\pm$0.9} & 79.9$\pm$0.3 &  \\ \midrule
\multirow{3}{*}{Citeseer}
 & 0.9 & 54.4$\pm$4.4 & 57.1$\pm$1.5 & 52.4$\pm$2.8 & 52.2$\pm$0.4 & 70.5$\pm$1.2 & 66.3$\pm$2.4 & 71.5$\pm$0.9 & 66.1$\pm$3.0 & 70.2$\pm$1.1 & \underline{72.8$\pm$0.3} & 72.3$\pm$0.3 & 72.5$\pm$0.5 & \textbf{76.8$\pm$0.4} & \multirow{3}{*}{71.7$\pm$0.4} \\
 & 1.8 & 64.2$\pm$1.7 & 66.7$\pm$1.0 & 64.3$\pm$1.0 & 59.0$\pm$0.5 & 70.6$\pm$0.9 & 69.0$\pm$1.1 & 71.2$\pm$0.7 & 69.2$\pm$1.2 & 71.7$\pm$0.9 & 71.7$\pm$0.2 & 72.6$\pm$0.6 & \underline{73.1$\pm$0.2} & \textbf{76.5$\pm$2.5} &  \\
 & 3.6 & 69.1$\pm$0.1 & 69.0$\pm$0.1 & 69.1$\pm$0.1 & 65.3$\pm$0.5 & 69.8$\pm$1.4 & 70.8$\pm$0.4 & 70.9$\pm$1.2 & 71.0$\pm$0.6 & \underline{74.0$\pm$0.4} & 72.5$\pm$0.5 & 72.6$\pm$0.5 & 71.5$\pm$0.3 & \textbf{76.7$\pm$0.2} &  \\ \midrule
\multirow{3}{*}{Pubmed}
 & 0.08 & 69.5$\pm$0.5 & 73.0$\pm$0.7 & 69.0$\pm$0.6 & 18.1$\pm$0.1 & 78.3$\pm$0.2 & 76.4$\pm$1.2 & 77.1$\pm$0.5 & \underline{78.9$\pm$0.7} & 77.1$\pm$1.1 & 77.1$\pm$0.3 & 77.7$\pm$0.7 & 77.3$\pm$0.1 & \textbf{79.9$\pm$0.2} & \multirow{3}{*}{77.1$\pm$0.3} \\
 & 0.15 & 73.8$\pm$0.8 & 75.4$\pm$0.7 & 73.7$\pm$0.8 & 28.7$\pm$4.1 & 77.1$\pm$0.3 & 77.5$\pm$0.4 & 78.0$\pm$0.3 & \underline{79.3$\pm$0.3} & 77.5$\pm$1.9 & 76.8$\pm$0.6 & 78.4$\pm$1.8 & 76.0$\pm$0.5 & \textbf{79.4$\pm$2.8} &  \\
 & 0.3 & 77.9$\pm$0.4 & 77.9$\pm$0.4 & 77.8$\pm$0.5 & 42.8$\pm$4.1 & 78.4$\pm$0.3 & 77.9$\pm$0.3 & 77.5$\pm$0.5 & \underline{79.4$\pm$0.3} & 78.0$\pm$0.5 & 78.1$\pm$0.3 & 78.2$\pm$0.8 & 77.8$\pm$0.2 & \textbf{80.0$\pm$1.0} &  \\ \midrule
\multirow{3}{*}{Ogbn-arxiv}
 & 0.05 & 47.1$\pm$3.9 & 52.4$\pm$1.8 & 47.2$\pm$3.0 & 35.4$\pm$0.3 & 59.2$\pm$1.1 & 59.0$\pm$1.8 & 59.6$\pm$0.5 & 58.7$\pm$1.7 & 60.6$\pm$1.1 & \underline{63.8$\pm$0.3} & 63.7$\pm$0.8 & \textbf{64.1$\pm$0.4} & 58.2$\pm$1.7 & \multirow{3}{*}{71.3$\pm$0.1} \\
 & 0.25 & 57.3$\pm$1.1 & 58.6$\pm$1.2 & 56.8$\pm$0.8 & 43.5$\pm$0.2 & 63.2$\pm$0.3 & 64.6$\pm$0.3 & 61.7$\pm$0.3 & 64.2$\pm$0.5 & 65.4$\pm$0.8 & \textbf{66.7$\pm$0.4} & 63.8$\pm$0.6 & \underline{66.2$\pm$0.1} & 57.9$\pm$2.1 &  \\
 & 0.5 & 60.0$\pm$0.9 & 60.4$\pm$0.8 & 60.3$\pm$0.4 & 50.4$\pm$0.1 & 64.0$\pm$0.4 & 65.2$\pm$0.8 & 58.7$\pm$0.6 & 65.1$\pm$0.7 & 65.9$\pm$0.6 & \textbf{67.6$\pm$0.3} & 64.1$\pm$0.3 & \underline{67.0$\pm$0.2} & 57.8$\pm$0.7 &  \\ \midrule
\multirow{3}{*}{Flickr}
 & 0.1 & 41.8$\pm$2.0 & 42.5$\pm$1.8 & 42.0$\pm$0.7 & 41.9$\pm$0.2 & 46.5$\pm$0.4 & 45.5$\pm$0.8 & 46.1$\pm$0.3 & 45.4$\pm$0.4 & 46.6$\pm$0.3 & 44.8$\pm$0.5 & \textbf{49.9$\pm$0.8} & 47.1$\pm$0.1 & \underline{48.4$\pm$0.8} & \multirow{3}{*}{47.1$\pm$0.1} \\
 & 0.5 & 44.0$\pm$0.4 & 43.9$\pm$0.9 & 43.2$\pm$0.1 & 44.5$\pm$0.1 & 47.1$\pm$0.1 & 46.0$\pm$0.4 & 45.9$\pm$0.4 & 46.0$\pm$0.4 & 46.6$\pm$0.2 & 46.4$\pm$0.2 & \textbf{49.4$\pm$1.3} & 47.1$\pm$0.0 & \underline{48.1$\pm$1.4} &  \\
 & 1 & 44.6$\pm$0.2 & 44.4$\pm$0.6 & 44.1$\pm$0.4 & 44.6$\pm$0.1 & 47.1$\pm$0.1 & 46.1$\pm$0.3 & 46.4$\pm$0.2 & 46.2$\pm$0.4 & 46.8$\pm$0.2 & 46.7$\pm$0.2 & \textbf{49.9$\pm$0.6} & 46.8$\pm$0.1 & \underline{48.0$\pm$0.9} &  \\ \midrule
\multirow{3}{*}{Reddit}
 & 0.05 & 46.1$\pm$4.4 & 53.1$\pm$2.5 & 46.6$\pm$2.3 & 40.9$\pm$0.5 & 88.0$\pm$1.8 & 80.0$\pm$3.1 & 84.2$\pm$0.7 & OOM & 90.5$\pm$0.2 & 91.2$\pm$0.1 & \textbf{92.9$\pm$0.3} & 90.7$\pm$0.1 & \underline{91.3$\pm$0.1} & \multirow{3}{*}{94.1$\pm$0.0} \\
 & 0.1 & 58.0$\pm$2.2 & 62.7$\pm$1.0 & 53.0$\pm$3.3 & 42.8$\pm$0.8 & 89.6$\pm$0.7 & 84.6$\pm$1.6 & 80.6$\pm$0.4 & OOM & 91.2$\pm$0.2 & \underline{92.4$\pm$0.0} & \textbf{93.1$\pm$0.2} & 91.0$\pm$0.1 & 90.1$\pm$0.9 &  \\
 & 0.2 & 66.3$\pm$1.9 & 71.0$\pm$1.6 & 58.5$\pm$2.1 & 47.4$\pm$0.9 & 90.1$\pm$0.5 & 87.9$\pm$1.2 & 84.1$\pm$0.3 & OOM & 92.2$\pm$0.1 & \underline{92.7$\pm$0.1} & \textbf{93.2$\pm$0.4} & 90.5$\pm$0.0 & 90.8$\pm$0.5 &  \\ \bottomrule
\end{tabular}
\end{table*}

\noindent \textbf{Hyperparameter Settings.} In the condensation step of \method, we considered the decomposition converged either after 200 epochs or when the absolute difference in reconstruction error between epochs $t$ and $t+1$ is below $10^{-7}$. Following~\citet{gcond}, during evaluation, we train a 2-layer GCN with 256 hidden units and no dropout for 600 epochs on the condensed graph, selecting the model with the lowest validation loss for performance evaluation on the original test set. Moreover, we reported the average accuracy and standard deviations of ten runs (condensation and evaluation steps).

To optimize the core tensor $\T{R}$ and the factor matrix $\mathbf{U}$, we used the Adam optimizer~\cite{Adam}. We tuned the learning rates for the reconstruction and evaluation steps within \{0.1, 0.01, 0.001, 0.0001\}, and the weight decay in a range of \{0, 0.01, 0.001, 0.0001\}. Additionally, the random edge additions and removals used to generate each augmented view of the original adjacency matrix are tuned across \{0.05, 0.1, 0.15, 0.2\}. Lastly, the number of views was tuned in a range of \{1, 2, 3, 4, 5\}, where the value 1 corresponds to a single-view decomposition. All hyperparameter optimization is performed using the Bayesian optimizer provided by Weights and Biases~\cite{wandb}.

\begin{figure*}[!htbp]
    \centering
    \includegraphics[width=\linewidth]{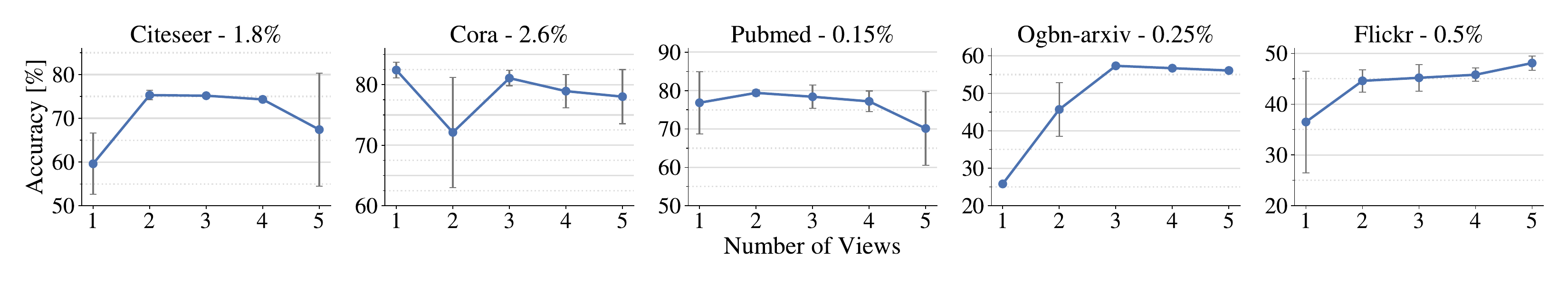}
    \Description{}\vspace{-0.35in}
    \caption{Accuracy scores achieved by our proposed method on graphs with varying numbers of views. The values following each dataset name represent the condensation ratio applied in this ablation study. The experiments were run ten times and we report the average accuracy and the respective error bars.}
    \label{fig:ablation_views}
\end{figure*}

\subsection{Experimental Results}\label{sec:experimental_results}
\noindent \textbf{Comparison with baselines.} To begin our analysis, we evaluate the performance of our method against selected baselines across six datasets, using three condensation ratios commonly utilized in the literature. We present the average performance across ten runs and the corresponding standard deviation in Table~\ref{tab:results}. Notably, our method yields superior performance on the Citeseer, Cora, and Pubmed datasets. For example, it achieves a 4.0\%, 3.4\%, and 2.7\% improvement in accuracy across the three condensation ratios applied to Citeseer. Moreover, \method exhibits lossless performance in all settings for Citeseer, Pubmed, and Flickr, as well as in the 1.3\% and 2.6\% ratios for Cora.

Analyzing the results on Flickr, a challenging dataset in our evaluation, we observe that \method performs robustly, securing the second-best results across all condensation ratios and achieving lossless performance. Similarly, \method performs competitively on Reddit, consistently ranking among the top-performing methods. For instance, at the 0.05\% condensation ratio, it achieves 91.3\% accuracy (the second-best result overall) and maintains comparable performance at higher ratios. These findings highlight the effectiveness of our method in transductive and inductive settings, as well as in large datasets (i.e., Flickr and Reddit). Although results on Ogbn-arxiv do not surpass the strongest baselines, \method maintains competitive performance and demonstrates robustness across diverse datasets, indicating potential for further development. \\

\noindent \textbf{On the number of graph views.} Next, we present an experiment we conducted to analyze the impact of the number of augmented views used in condensing the original graph. In this evaluation, we assessed our method on graphs with varying counts, ranging from 1 to 5. It is important to note that the single-view case corresponds to our first attempt, discussed in Section~\ref{sec:single_view}. Figure~\ref{fig:ablation_views} shows how the number of views influences overall performance.

Our results indicate that, except for Cora, using a multi-view augmented graph yields better performance than the single-view one. For example, on the Ogbn-arxiv dataset, we observed a 31.5\% performance improvement when using three views instead of just one, highlighting the effectiveness of using augmented views.\\

\begin{table}[!htp]
\caption{Accuracy (\%) of different GNN architectures on the graphs condensed by our method, GCond, and SFGC. The reported values are an average of ten runs. Best and second-best averages are in bold and underlined, respectively.}
\label{tab:different_gnns}
\setlength{\tabcolsep}{5pt}
\footnotesize
\centering
\begin{tabular}{lccccccc}
\toprule
\multirow{2}{*}{Datasets (r\%)} & \multirow{2}{*}{Methods} & \multicolumn{5}{c}{Models} & \multirow{2}{*}{Avg.} \\ \cmidrule(lr){3-7}
& & GCN & APPNP & Cheby & SGC & SAGE & \\ \midrule
\multirow{4}{*}{Cora (2.6)}
& GCond & 80.1 & 78.5 & 76.0 & 79.3 & 78.4 & 78.5 \\
& SFGC  & 79.3 & 78.8 & 79.0 & 79.1 & 80.0 & \underline{79.2} \\
& GCDM  & 80.5 & 79.7 & 75.5 & 77.7 & 77.5 & 78.2 \\
& GCTD  & 84.0 & 85.5 & 76.2 & 82.5 & 74.6 & \textbf{80.5} \\ \midrule
\multirow{4}{*}{Citeseer (1.8)}
& GCond & 70.6 & 69.6 & 68.3 & 70.3 & 66.2 & 69.0 \\
& SFGC  & 69.0 & 70.5 & 71.8 & 71.8 & 71.7 & 71.0 \\
& GCDM  & 71.7 & 73.6 & 66.1 & 73.6 & 71.1 & \underline{71.2} \\
& GCTD  & 76.5 & 73.8 & 75.2 & 77.4 & 73.5 & \textbf{75.3} \\ \midrule
\multirow{4}{*}{Pubmed (0.08)}
& GCond & 77.1 & 76.8 & 75.9 & 77.1 & 76.9 & 76.8\\
& SFGC  & 77.5 & 76.3 & 77.7 & 77.8 & 76.3 & \underline{77.1}\\
& GCDM  & 77.1 & 78.3 & 71.5 & 74.9 & 77.0 & 75.8 \\
& GCTD  & 79.9 & 79.2 & 78.0 & 79.8 & 79.0 & \textbf{79.2} \\ \midrule
\multirow{4}{*}{Flickr (0.5)}
& GCond & 47.1 & 45.9 & 42.8 & 46.1 & 46.2 & \underline{45.6} \\
& SFGC  & 46.0 & 40.7 & 45.4 & 42.5 & 47.0 & 44.3 \\
& GCDM  & 46.4 & 46.0 & 42.4 & 45.8 & 42.6 & 44.6 \\
& GCTD  & 48.1 & 46.9 & 42.7 & 45.4 & 47.2 & \textbf{46.1} \\ \bottomrule
\end{tabular}
\end{table}

\noindent \textbf{Performance across different GNNs.} We also evaluated the performance of various GNN architectures on the condensed graphs generated by \method and compared them to GCond, SFGC, and GCDM, which represent diverse approaches to graph condensation. Consistent with the evaluation protocol of \citet{gcond} and other follow-up works, we trained 2-layer versions of APPNP~\cite{appnp}, ChebyNet~\cite{chebynet}, SGC~\cite{Wu2019SimplifyingGCN}, and GraphSAGE~\cite{Hamilton2017GraphSAGE} using the synthesized graphs and measured their performance on the original graph’s test set. For this evaluation, we selected datasets across three size categories: Citeseer and Cora (small), Pubmed (medium), and Flickr (large), with condensation ratios of 1.8\%, 2.6\%, 0.15\%, and 0.5\%, respectively. We report the average accuracy over ten runs in Table~\ref{tab:different_gnns}.

Focusing on GCTD, we observe that APPNP, SGC, and GCN generally achieve strong results, often surpassing the average performance across datasets. While APPNP performs best on Cora and SGC leads on Citeseer and Pubmed, GCN consistently stays competitive. GraphSAGE also stands out on Flickr, achieving the highest score among the GNNs for that dataset. In contrast, ChebyNet underperforms, with below-average results on most datasets.

Analyzing the baselines, we observe that GCTD outperforms all methods in terms of average accuracy across datasets. For instance, on Citeseer, GCTD improves over GCDM (the second-best method) by 4.1\%. On Pubmed, it surpasses SFGC by 2.1\%, and on Cora, it leads by 1.3\%. Even on Flickr, where all methods struggle, GCTD still achieves the best average, showing a 0.5\% gain over GCond.

\begin{figure}[!tbh]
    \centering
    \includegraphics[width=\linewidth]{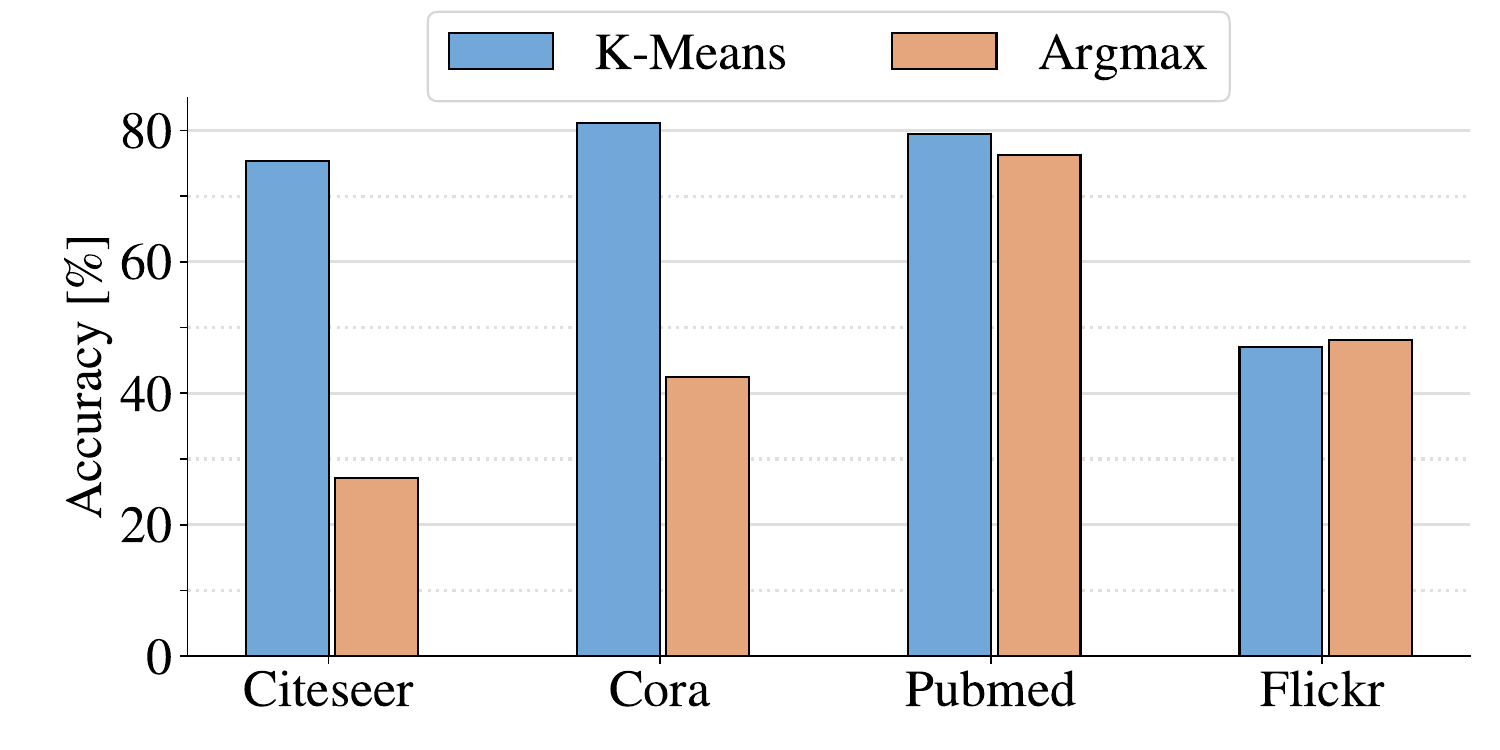}
    \Description{Accuracy of \method with K-Means and Argmax as the method employed to compute the synthetic node assignments from factor matrix $\mathbf{U}$. In this ablation study, we used Citeseer, Cora, Pubmed, and Flickr with the condensation ratio set to 1.8\%, 2.6\%, 0.15\%, and 0.5\%, respectively.}
    \vspace{-0.1in}
    \caption{Accuracy of \method with K-Means and Argmax as the method employed to compute the synthetic node assignments from factor matrix $\mathbf{U}$. In this ablation study, we used Citeseer, Cora, Pubmed, and Flickr with the condensation ratio set to 1.8\%, 2.6\%, 0.15\%, and 0.5\%, respectively.}
    \label{fig:ablation_kmeans}
\end{figure}

\noindent \textbf{Ablation on the synthetic node assignment.} 
We conducted an ablation study to evaluate the effectiveness of the synthetic node assignment method we leveraged. Specifically, we used the \emph{argmax} operation on each row of the factor $\mathbf{U}$, assigning the selected node to the corresponding synthetic node. We then compared the performance of this approach to that of K-Means, the default method in \method. In this ablation, we follow the same settings used in the previous experiment, employing Citeseer, Cora, Pubmed, and Flickr with 1.8\%, 2.6\%, 0.15\%, and 0.5\% condensation ratios, respectively. 

As shown in Figure~\ref{fig:ablation_kmeans}, while argmax provides a straightforward method for generating synthetic nodes, its simplicity can sometimes limit its ability to produce high-quality synthetic nodes. Specifically, we can observe that only selecting the maximum value in each row as the cluster membership is insufficient to capture the necessary information for the condensed graph. On the other hand, even though KMeans is computationally complex compared to argmax, it offers greater performance. It typically obtains better clustering of nodes by effectively grouping them into synthetic nodes based on the optimized factor matrix. This matrix, derived during the decomposition process, captures co-clustered nodes exhibiting similar patterns, leading to more meaningful synthetic representations~\cite{smacd}.

\begin{table}[!tbh]
\centering
\caption{Running time analysis of the proposed method and the SGC version of the baselines. The experiments were performed on a single NVIDIA RTX A6000. OOM occurred at 49GB of memory usage.}
\label{tab:running_time}
\setlength{\tabcolsep}{4.2pt}
\small
\begin{tabular}{lcccccc}
    \toprule
    Dataset (r\%) & GCond & SGDD & SFGC & GCDM & SNTK & GCTD \\ \midrule
    Citeseer (0.9) & 70.8 & 140.9 & 63.0 & 131.3 & 2.9 & 32.1 \\
    Cora (1.3) & 90.3 & 115.1 & 62.4 & 174.7 & 3.1 & 42.8 \\
    Pubmed (0.08) & 57.1 & 2200.5 & 61.6 & 75.4 & 4.2 & 48.1 \\
    Ogbn-arxiv (0.05) & 725.0 & 2073.3 & 238.0 & 480.0 & 925.4 & 965.6 \\
    Flickr (0.1) & 470.4 & 361.5 & 42.9 & 225.0 & 148.7 & 222.9 \\
    Reddit (0.05) & 3780 & 10350 & 2420 & 1617.1 & OOM & 2515.3 \\\bottomrule 
\end{tabular}
\end{table}

\begin{table*}[!tbh]
\centering
\caption{Comparison between the condensed graphs generated by GCTD and original graphs.}
\small
\label{tab:comparison_original}
\setlength{\tabcolsep}{10.4pt}
\begin{tabular}{@{}l cc cc cc cc cc@{}}
\toprule
          & \multicolumn{2}{c|}{Citeseer (0.9\%)} & \multicolumn{2}{c|}{Cora (1.3\%)} & \multicolumn{2}{c|}{Pubmed (0.08\%)} & \multicolumn{2}{c|}{Flickr (0.1\%)} & \multicolumn{2}{c}{Ogbn-arxiv (0.05\%)} \\ 
          \cmidrule{2-3} \cmidrule{4-5} \cmidrule{6-7} \cmidrule{8-9} \cmidrule{10-11}
          & Whole              & \method            & Whole            & \method          & Whole                 & \method            & Whole             & \method           & Whole               & \method         \\ \midrule
Accuracy  & 71.7               & 76.8               & 81.4             & 81.4             & 77.1                  & 79.9               & 47.1              & 48.4              & 71.3                & 58.2            \\
\#Nodes   & 3,327              & 30                 & 2,708            & 35               & 19,717                & 15                 & 44,625            & 44                & 169,343             & 84              \\
\#Edges   & 4,732              & 93                 & 5,429            & 72               & 44,338                & 67                 & 218,140           & 990               & 1,116,243           & 37               \\
Sparsity  & 0.09\%             & 22.91\%            & 0.15\%           & 12.1\%           & 0.01\%                & 63.8\%             & 0.02\%            & 100\%             & 0.09\%              & 1.1\%            \\
Storage   & 47.1 MB            & 0.51 MB            & 14.9 MB          & 0.23 MB          & 40 MB                  & 0.05 MB            & 86.8 MB           & 0.17 MB           & 100.4 MB            & 0.08 MB          \\ \bottomrule
\end{tabular}
\vspace{0.2in}
\end{table*}

\noindent \textbf{Running time.}\label{app:running_time}
In Table~\ref{tab:running_time}, we compare the running time of our method against GCond, SGDD, SFGC, GCDM, and SNTK. To ensure a fair comparison, we adopted the SGC implementation for all baselines, as it offers faster runtime than their GCN counterparts. Overall, our method strikes a balance between speed and performance. It consistently outperforms SGDD across all datasets, significantly reducing condensation time. For instance, on Citeseer, our method is over four times faster than SGDD and twice as fast as SFGC, one of the fastest baseline models. On larger datasets our method remains considerably faster than SGDD.

When compared to GCond and GCDM, our method shows better efficiency on smaller datasets and Flickr, while also outperforming GCond on Reddit. It is, however, slightly slower than GCDM on Reddit. Notably, SFGC remains among the fastest methods due to its structure-free condensation approach, which aligns with the findings of~\citet{mlpinit} that matrix multiplication with the adjacency matrix is the most time-consuming operation in GNNs. Lastly, while SNTK performs well on smaller graphs, its runtime advantage diminishes as graph size increases. On Reddit, it runs out of memory (49GB in our setup) due to its expensive kernel computation. These results highlight that our method scales well across diverse datasets, delivering competitive performance in graph condensation without incurring excessive computational costs.\\

\begin{figure}[!tbh]
    \centering
    \subfigure[Citeseer - 0.9\%]{ \label{fig:cite_graph}
        \includegraphics[width=0.22\textwidth]{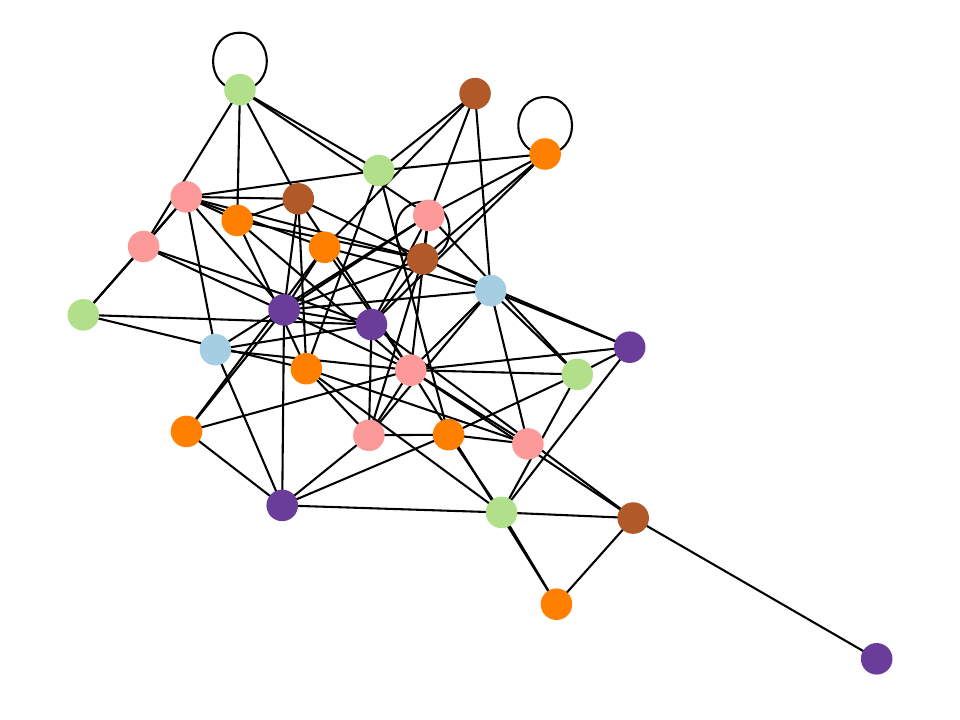}
    } \vspace{-0.1in}
    \subfigure[Cora - 1.3\%]{ \label{fig:graph_cora}
        \includegraphics[width=0.22\textwidth]{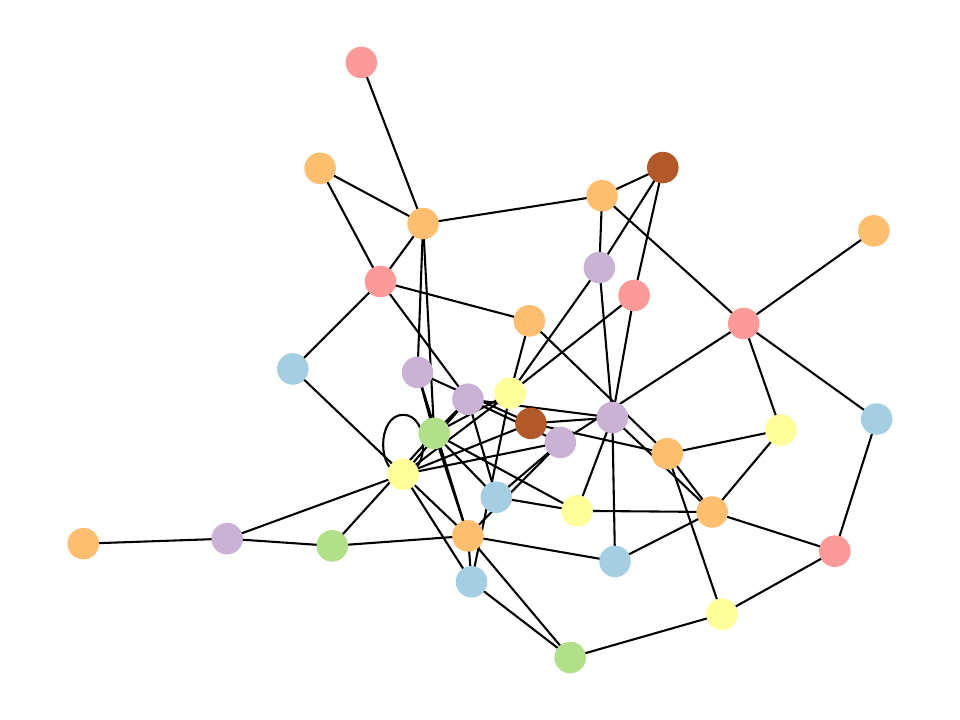}
    }
    \subfigure[Pubmed - 0.08\%]{ \label{fig:graph_pubmed}
        \includegraphics[width=0.22\textwidth]{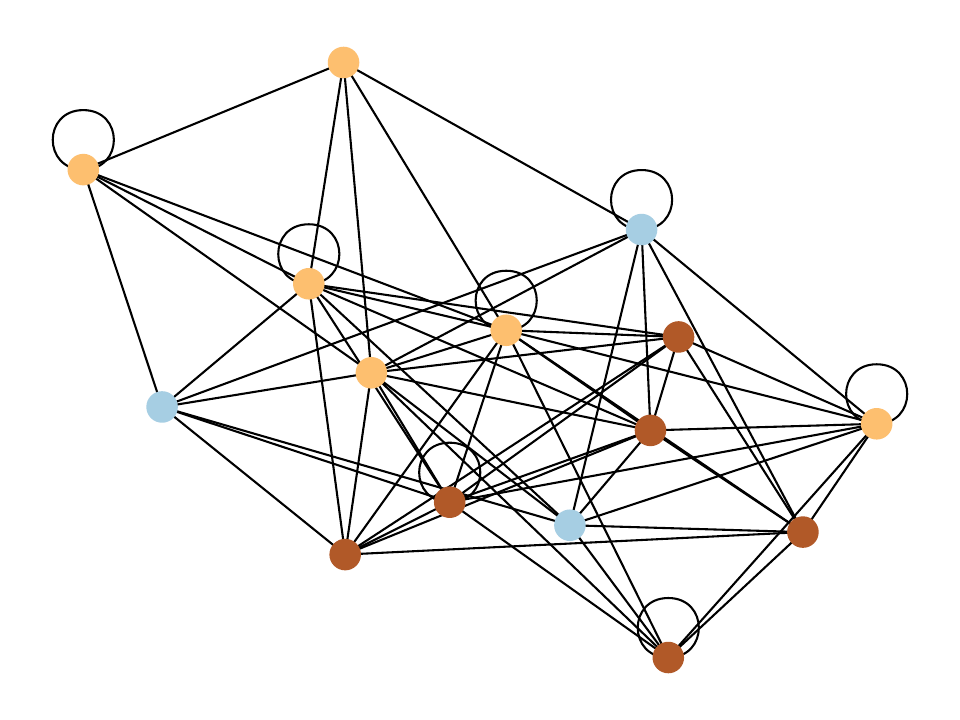}
    }   \vspace{-0.1in}
    \subfigure[Flickr - 0.1\%]{ \label{fig:graph_flickr}
        \includegraphics[width=0.22\textwidth]{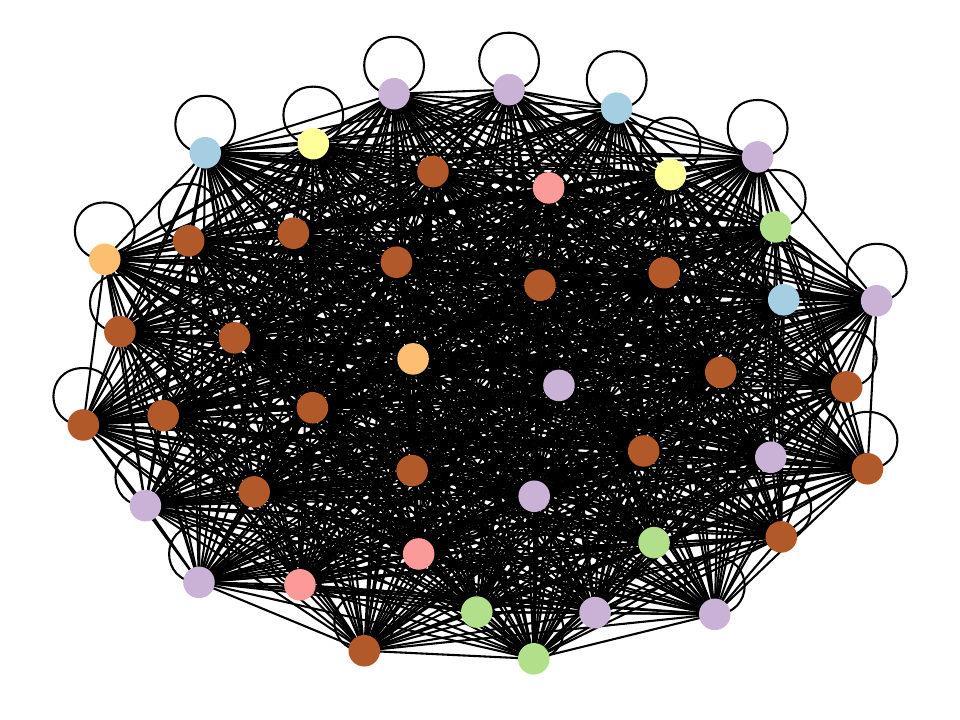}
    }
    \caption{Visualization of condensed graphs generated by \method. Each node represents a synthetic node, with its color indicating the corresponding class.}
    \label{fig:graphs}
\end{figure}

\noindent \textbf{Visualization of the condensed graphs.} We visualize the condensed graphs generated with the smallest condensation ratio for Citeseer, Cora, Pubmed, and Flickr in Figure~\ref{fig:graphs}. Several key observations emerge from these visualizations. First, \method produces graphs with well-defined structures and introduces self-loops for certain nodes. Additionally, it reduces the homophily present in the original graphs, indicating a diminished reliance on this property. This is particularly intriguing, given that the GNNs we used typically depend on graph homophily for optimal performance, as discussed in prior work~\cite{homophily_paper}. Finally, \method generates a complete graph for Flickr, implying that the GNN now relies on information from every node to compute individual node representations. As a result, the GNN relies on the weights and features of nodes to differentiate between information from various neighbors. \\

\noindent \textbf{Comparison with original graphs.} Table~\ref{tab:comparison_original} presents a comparison of various properties between the original graphs and the condensed graphs generated by GCTD. The results show that \method achieves comparable performance on Citeseer, Cora, Pubmed, and Flickr, while significantly reducing the number of nodes and edges, as well as requiring less storage. Furthermore, the condensed graphs are denser than their original counterparts, and a notable behavior is observed on Flickr, where the learned graph is complete.

%% file: content/7-conclusion.tex
\section{Conclusion}
In this paper, we introduced \method, a novel framework for graph condensation through the decomposition of multi-view augmented graphs. Moreover, we also showed the extent to which tensor decomposition methods can generate smaller graphs while maintaining the performance of GNNs on downstream tasks. We evaluated \method on six real-world datasets spanning transductive and inductive learning settings. Additionally, we conducted an in-depth analysis to demonstrate the effectiveness of our approach.

Our results indicate that \method outperforms existing methods on three of the six datasets and achieves competitive performance on the larger datasets. Furthermore, our proposed method performs well on transductive and inductive datasets. For future work, we plan to explore different decomposition techniques to generate condensed graphs. In addition, we will explore different ways of performing augmentation to improve the quality of the synthesized graph. Lastly, we plan to investigate whether replacing the current hard assignment with a soft membership approach can improve the effectiveness of \method.